\documentclass[]{article}
\usepackage{proceed2e}
\usepackage{times}
\usepackage{amsmath,amssymb,amsthm,amsfonts,dsfont}
\usepackage{mathtools}
\usepackage{color}
\usepackage[usenames,dvipsnames]{xcolor}
\usepackage{verbatim}
\usepackage{url}
\usepackage{array}
\usepackage[round]{natbib}
\newcommand{\cI}{{\mathcal I}}
\newcommand{\cJ}{{\mathcal J}}
\newtheorem*{theorem*}{Theorem}
\newtheorem*{lemma*}{Lemma}
\newtheorem{lemma}{Lemma}
\newtheorem{theorem}{Theorem}
\newtheorem{corollary}{Corollary}

\title{Novel Bernstein-like Concentration Inequalities for the Missing Mass}

\small
\author{
\begin{tabular}{*{2}{>{\centering}p{.5\textwidth}}}
\large Bahman Yari Saeed Khanloo & \large Gholamreza Haffari \tabularnewline
Monash University & Monash University \tabularnewline
\url{bahman.khanloo@monash.edu} & \url{gholamreza.haffari@monash.edu} 
\end{tabular}
}

\date{}

\begin{document}
\maketitle
\begin{abstract}
We are concerned with obtaining novel concentration inequalities for the {\em missing mass}, i.e.
the total probability mass of the outcomes not observed in the sample. 
We not only derive - for the first time - distribution-free Bernstein-like \linebreak deviation bounds with 
{\em sublinear} exponents in  deviation size for
missing mass, but also improve the results of \cite{McAllester2003} and \cite{Berend_Kontorovich_Bound, Berend2012} for small 
deviations which is the most interesting case in learning theory. 
It is known that the majority of standard inequalities cannot be directly used to \linebreak analyze heterogeneous 
 sums i.e. sums whose terms have 
large difference in magnitude. Our generic and intuitive approach shows that the \linebreak heterogeneity issue 
introduced in \cite{McAllester2003} is resolvable at least in the case of 
missing mass via regulating the terms using our novel thresholding technique.
\end{abstract}

\section{INTRODUCTION}
\nocite{pemantle-na}
Missing mass is the total probability associated 
to the \linebreak outcomes that have not been seen in the sample which is one of the important 
quantities in machine learning and statistics. It connects density 
estimates obtained from a given sample to the population for discrete distributions: the less the missing mass, the more useful the 
information that can be extracted from the dataset.
Roughly speaking, the more the missing mass is the less we can discover about the true unknown underlying distribution which would imply  
the less we can statistically generalize to the whole population. In other words, missing mass measures how representative a 
given dataset is assuming that it has been sampled according to the true distribution.

Often, one is interested in understanding the behaviour 
of the missing mass as a random variable.
One of the \linebreak important approaches in such studies involves bounding the fluctuations of the random variable 
around a certain \linebreak quantity namely its mean.  
Concentration inequalities are powerful tools for performing analysis of this type.
Let $X$ be any non-negative real-valued random variable with finite mean.
The goal is to establish for any $\epsilon>0$, probability bounds of the form
\begin{align}
\mathbb{P}(X - \mathbb{E}[X] \leq -\epsilon) \leq
\exp(-\eta_l (\epsilon)), \nonumber \\
\mathbb{P}(X - \mathbb{E}[X] \geq \epsilon) \leq \exp(-\eta_u (\epsilon)), \label{eq:bernoullib1}
\end{align}
where $\eta_l(\epsilon)$ and $\eta_u(\epsilon)$ are some non-decreasing functions of $\epsilon$ and where it is desirable to find the largest
such functions for variable $X$ and for the `target' interval of $\epsilon$.  These bounds are commonly called lower and upper deviations bounds respectively. 
In most practical scenarios, we are in a non-asymptotic setting where we have access to a sample $X_1,...,X_n$ and we would 
like to derive concentration inequalities
that explicitly describe dependence on sample size $n$.
Namely, we would like to obtain bounds of the form
\begin{align}
\mathbb{P}(X - \mathbb{E}[X] \leq -\epsilon) \leq
\exp(-\eta_l (\epsilon,n)), \nonumber \\
\mathbb{P}(X - \mathbb{E}[X] \geq \epsilon) \leq 
\exp(-\eta_u (\epsilon,n)), \label{eq:bernoullib}
\end{align}
where $\eta_l (\epsilon,n)$ and $\eta_u (\epsilon,n)$ are both non-decreasing functions of $\epsilon$ and $n$.
Many of such bounds are distribution-free i.e. they hold irrespective of the underlying distribution.

\cite{tg_rate} established concentration inequalities for the missing mass for the first time. 
A follow-up work by \cite{McAllester2003} pointed out inadequacy of standard inequalities, 
developed a thermodynamical viewpoint for addressing this issue and sharpened these bounds. 
\cite{Berend_Kontorovich_Bound} further refined the bounds via arguments similar to 
Kearns-Saul inequality (\cite{KearnsS98}) and logarithmic Sobolev inequality (\cite{boucheron2013concentration}). 
These \linebreak previous works, however, not only 
involve overly specific 
approaches to concentration and handling heterogeneity \linebreak issue
but also do not yield sharp bounds for small deviations which 
is the most interesting case in learning theory.

In this paper, we shall derive distribution-free concentration inequalities for missing mass in a novel way.
The \linebreak primary objective of our approach is to introduce a notion of {\em heterogeneity control} which allows us to
{\em regulate} the magnitude of bins in histogram of the discrete distribution being analyzed.
This mechanism in turn enables us to control the behaviour of central quantities such as the variance or martingle differences 
of the random variable in question.
These are the main quantities that appear in standard concentration inequalities such as Bernstein, Bennett and McDiarmid just 
to name a few.
Consequently, instead of discovering a new method for bounding fluctuations of each random variable of interest, we will be able 
to directly apply standard 
inequalities to obtain probabilistic bounds on many discrete random variables
including missing mass. 

The rest of the paper is structured as follows. Section \ref{sec:defs} contains the background information and introduces the notations. 
Section \ref{sec:main_res} outlines motivations and the main contributions. 
In Section \ref{neg_dep-info_monot}, we explain negative dependence, information monotonicity and develop a few fundamental tools 
whereas Section \ref{sec:proofs} presents the proofs of our upper and lower deviation
bounds based on these tools. Finally, Section \ref{sec:conclusion} concludes the paper and compares our 
bounds with existing results for small deviations.

\section{PRELIMINARIES}
\label{sec:defs}
In this section, we will provide definitions, notations and and other background material.
\\
Consider $P: \cI \rightarrow [0,1]$ to be a fixed but unknown 
discrete distribution on some finite or countable non-empty set $\cI$ with $|\cI|=N$. 
Let $\{ w_i: i \in \cI \}$ be the probability (or frequency) of drawing the $i$-th outcome.
Moreover, suppose that we observe an i.i.d. sample $\{X_j\}_{j=1}^n$ from this distribution with $n$ being the sample size.
Now, missing mass is defined as the total probability mass corresponding to the outcomes that are not present in our sample.
Namely, missing mass is a random variable that can be expressed as:
\begin{align}
\label{eq:bernoullisum_mm}
Y := \sum_{i \in \cI}  w_i Y_i,
\end{align}
where we define each $\{ Y_i: i \in \cI \}$ to be a Bernoulli \linebreak variable that takes on $0$ if the $i$-th outcome exists
in the sample and $1$ otherwise. Namely, we have
\begin{align}
 Y_i = \mathds{1}_{[  (X_1 \neq i) \land (X_2 \neq i) \land \cdots \land (X_n \neq i)  ]}.
\end{align}

We assume that for all $i \in \cI$, $w_i>0$ and $\sum_{i\in \cI} w_i = 1$. \linebreak
Denote $P(Y_i = 1) = q_i$ and $P(Y_i = 0) = 1-q_i$ and let us suppose that $Y_i$s are independent:
as we will see later in this section, such an assumption will not impose a burden on our proof structure and flow. 
Hence, we will have 
that $q_i=\mathbb{E}[Y_i]=(1-w_i)^n \leq e^{-n w_i}$
where $q_i \in (0,1)$.  Namely, defining $f:(1,n) \rightarrow (e^{-n}, \frac{1}{e}) \subset (0,1)$ 
where $f(\theta)=e^{-\theta}$  with $\theta \in D_f$ and taking 
$w_i > \frac{\theta}{n}$ amounts to  $q_i(w_i) \leq f(\theta)$.
This provides a basis for our `thresholding' technique that we will employ in our proof.

\noindent Choosing the representation (\ref{eq:bernoullisum_mm}) for missing mass, one has
\begin{align}
{\mathbb{E}[Y]}_{\cI}=\sum_{i \in \cI} w_i q_i = \sum_{i \in \cI} w_i(1-w_i)^n, \\
{V[Y]}_{\cI}=\sum_{i \in \cI} w_i^2 \text{\sc var\;}[Y_i], \label{variance_proxy}\\
{\underline{\sigma}^2}_{\cI} \coloneqq \sum_{i \in \cI} w_i \text{\sc var\;}[Y_i], \label{weighted_var}
\end{align}
where we have introduced the weighted variance notation ${\underline{\sigma}^2}$ and 
where each quantity is attached to a set over which it is defined.
Note that $\text{\sc var\;}[Y_i]$ is the individual variance corresponding to $Y_i$ which is defined as
\begin{align}
\text{\sc var\;}[Y_i] = q_i (1-q_i) = (1-w_i)^n \big(1-(1-w_i)^n \big).
\end{align}

\par One can define the above quantities not just over the set $\cI$ but on some (proper) subset of it that may depend 
on or be described by some variable(s) of interest.
For instance, in our proofs the variable $\theta$ may be responsible for choosing $\cI_{\theta} \subseteq \cI$ over which 
the above quantities will
be evaluated. For lower deviation and upper deviation, we
find it convenient to refer to the associated set by $\mathcal{L}$ and $\mathcal{U}$ respectively. 
Likewise, we will use subscripts $l$ and $u$ to refer to objects that characterize lower deviation and upper deviation 
\linebreak respectively. 
Also, we use the notation $Y^{ij}=Y_i,...,Y_j$ to refer to sequence of variables whose index starts at $i$-th variable and ends at
$j$-th variable. Finally, other notation or definitions may be introduced within the body of the proof when required.

We will encounter Lambert $W$-function - also known as product logarithm function - in this paper which describes the inverse relation of $f(x)=x e^x$ and which
cannot be expressed in terms of elementary functions. This function is double-valued when $x \in \mathbb{R}$. However, it 
becomes \linebreak invertible in restricted domain. 
The lower branch of it is \linebreak denoted by $W_{-1}(.)$, which is the only branch that will prove beneficial in this paper.
The reader is advised to refer to \cite{lambert} for a detailed treatment. 

Throughout the paper, we shall use the convention that \linebreak capital letters refer to random variables whereas 
lower case \linebreak letters correspond to 
realizations thereof.

We will utilize Bernstein's inequality 
in our derivation. Suitable representations of this result are outlined below without the proof.

\begin{theorem*} \ {\bf [Bernstein]} \label{thm:bernstein}
Let $Z_1, ..., Z_N$ be independent zero-mean random variables such that one has $|Z_i| \leq \alpha$ almost surely for 
all $i$. Then, using Bernstein's inequality \linebreak (\cite{Bernstein_ineq}) one obtains for all $\epsilon>0$:
\begin{equation}\label{eq:bernsteina}
\mathbb{P} ( \sum_{i=1}^{N} Z_i > \epsilon )
 \leq \exp \Big(-\frac{{\epsilon}^2}{2(V + \frac{1}{3} \alpha \epsilon)}\Big),
\end{equation}
where $V=\sum_{i=1}^{N}\mathbb{E}[{Z_i}^2]$. 
\end{theorem*}
\noindent Now, consider the sample mean 
$\bar{Z} = n^{-1}\sum_{i=1}^n Z_i$ and let $\bar{\sigma}^2$ be the sample variance, 
namely \linebreak $\bar{\sigma}^2 \coloneqq n^{-1} \sum_{i=1}^n \text{\sc var\;}[{Z_i}]=n^{-1} \sum_{i=1}^n \mathbb{E}[{Z_i}^2]$. 
So, using (\ref{eq:bernsteina})
with $n \cdot \epsilon$ in the role of
$\epsilon$, we get 
\begin{equation}\label{eq:bernsteinb}
\mathbb{P} ( \bar{Z} > \epsilon )
 \leq \exp \Big(-\frac{n {\epsilon}^2}{2(\bar{\sigma}^2 + \frac{1}{3} \alpha \epsilon)}\Big).
\end{equation}
If $Z_1, ..., Z_n$ are, moreover, not just independent but also
identically distributed, then $\bar{\sigma}^2$ is equal to $\sigma^2$ i.e.
the \linebreak variance of each $Z_i$.
The latter presentation makes explicit: (1) the exponential decay with $n$; (2) the fact
that for $\bar{\sigma}^2 \leq \epsilon$ we get a tail probability with
exponent of order $n \epsilon$ rather than $n \epsilon^2$ (\cite{Lugosi_Concentration,boucheron2013concentration}) 
which has the potential to yield stronger bounds for small $\epsilon$.

\section{MOTIVATIONS AND MAIN RESULTS}
\label{sec:main_res}
In this section, we motivate this work by pointing out the heterogeneity challenge and how we approach it. Our 
bounds also improve the functional form of
the exponent, which is of independent significance. In the final part of this section, we summarize our main results.

\subsection{The Challenge and the Remedy}
\cite{McAllester2003} point out that for highly \linebreak heterogeneous sums of the
form (\ref{eq:bernoullisum_mm}), the standard form of Bernstein's
inequality (\ref{eq:bernsteina}) does not lead to
concentration inequalities of form (\ref{eq:bernsteinb}): at least for the
upper deviation of the missing mass, (\ref{eq:bernsteina}) does not
imply any non-trivial bounds of the form (\ref{eq:bernoullib}).  The
reason is basically the fact that the $w_i$ can vary wildly: 
some can be of order $O(1/n)$, other
may be constants independent of $n$. For similar reasons, other
standard inequalities such as Bennett, Angluin-Valiant and
Hoeffding cannot be used to get bounds on the missing mass of the
form (\ref{eq:bernoullib}) either (\cite{McAllester2003}). 

Having pointed out the
deficiency of these standard inequalities, \cite{McAllester2003}
succeed in giving bounds of the form (\ref{eq:bernoullib}) on the
missing mass, for a function $\eta(\epsilon,n) \propto n \epsilon^2$, both
with a direct argument and using the Kearns-Saul inequality
(\cite{KearnsS98}).  \linebreak Recently, the constants appearing in the bounds
were \linebreak refined by \cite{Berend_Kontorovich_Bound}. The bounds proven by
\cite{McAllester2003} and \cite{Berend_Kontorovich_Bound} are qualitatively
similar to Hoeffding bounds for i.i.d. random variables: they do {\em not}
improve the functional form from $n \epsilon^2 $ to $n \epsilon$ for small variances.

This leaves open the question whether it is also possible to derive
bounds which are more reminiscent of the Bernstein bound for i.i.d. random
variables (\ref{eq:bernsteinb}) which does exploit variance. In
this paper, we show that the \linebreak answer is a qualified yes: we give
bounds that depend on \linebreak weighted variance $\underline{\sigma}^2$ 
 defined in (\ref{weighted_var}) rather than sample \linebreak variance $\bar{\sigma}^2$ as in (\ref{eq:bernsteinb}) which is tight
exactly in the \linebreak important case when $\underline{\sigma}^2$ is small, and in which the \linebreak denominator in (\ref{eq:bernsteinb})
is specified by a factor depending on $\epsilon$; in the special case of the missing
mass, this factor turns out to be logarithmic in $\epsilon$ and a free parameter $\gamma$ as it will become clear later.

We derive - using Bernstein's inequality - novel bounds on missing mass that take into 
account explicit variance \linebreak information with
more accurate scaling and demonstrate their superiority for small deviations.

\subsection{Main Results}
\noindent Consider the following functions
\begin{align}
\gamma_{\epsilon}=-2W_{-1} \big(-\frac{\epsilon}{2 \sqrt e} \big), \\
c(\epsilon)=\frac{3(\gamma_{\epsilon}-1)}{5 \gamma_{\epsilon}^2}. 
\end{align}
Let $Y$ denote the missing mass, $n$ the sample size and $\epsilon$ the deviation size. 
\begin{theorem}
\label{main_theorem}

For any $0<\epsilon<1$ and any $n \geq \lceil {\gamma}_{\epsilon} \rceil -1$, we obtain the following upper deviation bound
\begin{equation}\label{eq:mm_bound_ud}
\mathbb{P}(Y - {\mathbb E} [Y] \geq \epsilon )
\leq e^{-c(\epsilon) \cdot n \epsilon}.
\end{equation} 
\end{theorem}

\begin{theorem}
For any $0 < \epsilon < 1$ and any $n \geq \lceil {\gamma}_{\epsilon} \rceil -1$, we obtain the following lower deviation bound 
\begin{align}\label{eq:mm_bound_ld1}
\mathbb{P}(Y - {\mathbb E} [Y] \leq -\epsilon )
\leq e^{-c(\epsilon) \cdot n \epsilon}. 
\end{align}
\end{theorem}

\begin{corollary}
For any $0<\epsilon<1$ and any $n \geq \lceil {\gamma}_{\epsilon} \rceil -1$, using union bound we obtain the following deviation bound 
\begin{align}
\mathbb{P}(\lvert Y - \mathbb{E} [Y] \rvert \geq \epsilon) \leq 2 \ e^{-c(\epsilon) \cdot n \epsilon}.
\end{align}
\end{corollary}
The proof of the above theorems is provided in Section \ref{sec:proofs}.
However, let us develop a few tools in Section \ref{neg_dep-info_monot} which will be used later in our proofs.

\section{NEGATIVE DEPENDENCE AND INFORMATION MONOTONICITY}
\label{neg_dep-info_monot}
Probabilistic analysis of most random variables and specifically the derivation of the majority of probabilistic bounds 
rely on independence assumption between variables
which offers considerable simplification and convenience.
Many random variables including the missing mass, however, consist of random components that are not independent.

Fortunately, even in cases where independence does not hold, one can still use some standard tools and methods 
provided variables are dependent in specific ways.
The \linebreak following notions of dependence are among the common \linebreak ways that prove useful in these settings:
negative \linebreak association and negative regression.

\subsection{Negative Dependence and Chernoff's Exponential Moment Method}
Our proof involves variables with a specific type of \linebreak dependence known as negative association. 
One can \linebreak infer concentration of sums of negatively associated random variables 
from the concentration of sums of their \linebreak independent copies in certain situations.
In exponential moment method, this property allows us to treat such \linebreak variables as independent in the context of 
probability \linebreak inequalities as we shall elaborate later in this section.

In the sequel, we present negative association and \linebreak regression and supply tools that will be essential in proofs.

\noindent \textbf{Negative Association: }
Any real-valued random variables $X_1$ and $X_2$ are negatively associated if
\begin{align}
\mathbb{E}[X_1 X_2] \leq \mathbb{E}[X_1] \cdot \mathbb{E}[X_2].
\end{align}
More generally, a set of random variables $X_1,...,X_m$ are negatively associated if for any disjoint subsets $A$ 
and $B$ of the index set $\{1,...,m\}$, we have
\begin{align}
\mathbb{E}[X_i X_j] \leq \mathbb{E}[X_i] \cdot \mathbb{E}[X_j] \quad \text{for} \quad i \in A, \ j \in B.
\end{align}

\noindent \textbf{Stochastic Domination: }
Assume that $X$ and $Y$ are real-valued random variables. Then, $X$ is said to stochastically dominate $Y$ if 
for all $a$ in the range of $X$ and $Y$ 
we have
\begin{align}
P(X \geq a) \geq P(Y \geq a). \label{stochastic_domination}
\end{align}
We use the notation $X \succeq Y$ to reflect (\ref{stochastic_domination}) in short.

\noindent \textbf{Stochastic Monotonicity: }
A random variable $Y$ is stochastically non-decreasing in random variable $X$ if
\begin{align}
x_1 \leq x_2 \ \Longrightarrow \ P(Y|X=x_1) \leq P(Y|X=x_2). \label{stoch_inc}
\end{align}
Similarly, $Y$ is stochastically non-increasing in $X$ if
\begin{align}
x_1 \leq x_2 \ \Longrightarrow \ P(Y|X=x_1) \geq P(Y|X=x_2). \label{stoch_dec}
\end{align}
The notations $(Y|X=x_1) \preceq (Y|X=x_2)$ and \linebreak $(Y|X=x_1) \succeq (Y|X=x_2)$ represent the 
above \linebreak definitions using the notion of stochastic domination. Also, we will use shorthands $Y \uparrow X$ and $Y \downarrow X$ 
to refer to the relations described by (\ref{stoch_inc}) and (\ref{stoch_dec}) respectively.

\noindent \textbf{Negative Regression: }
Random variables $X$ and $Y$ have negative regression dependence relation if $X \downarrow Y$.

\cite{Dubhashi-Ranjan} as well as \cite{kumar_frank} summarize numerous notable properties of 
negative association and negative regression. Specifically, the former provides a proposition that indicates that 
Hoeffding-Chernoff bounds apply to sums of 
negatively associated random variables.
Further, \cite{McAllester2003} generalize these observations to essentially any 
concentration result derived based on the exponential moment method by
drawing a connection between deviation  probability of a discrete random variable and 
Chernoff's entropy of a related distribution. 

We provide a self-standing account by presenting the proof for some of these existing results as well as  
developing \linebreak several generic tools that are applicable beyond missing mass problem.
\begin{lemma} \ {\bf [Binary Stochastic Monotonicity]} \label{lemma:monotone_bernoulli} \upshape
Let $Y$ be a binary random variable (Bernoulli) and let $X$ take on \linebreak values 
in a totally ordered set $\mathcal{X}$. Then, one has
\begin{align}
Y \downarrow X \ \Longrightarrow \ X \downarrow Y.
\end{align}
\begin{proof}
 For any $x$, we have
\begin{align}
P(Y=1 | \ X \leq x) &\geq \inf_{a \leq x} P(Y=1 | \ X=a) \nonumber \\
		    &\geq \sup_{a > x} P(Y=1 | \ X=a) \nonumber \\
		    &\geq P(Y=1 | \ X>x).
\end{align}
The above argument implies that random variables $Y$ and $\mathbf{1}_{X>x}$ are negatively associated and since 
the \linebreak expression  
$P(X>x| \ Y=1) \leq P(X>x | \ Y=0)$ holds for all $x \in \mathcal{X}$, it follows that 
$X \downarrow Y$.
\end{proof}
\end{lemma}

\begin{lemma} \ {\bf [Independent Binary Negative Regression]} \label{lemma:indep_bin_neg_reg} \upshape
Let $X_1,...,X_m$ be negatively associated random variables and $Y_1,...,Y_m$ be binary random 
variables (Bernoulli) such that either $Y_i \downarrow X_i$ or $Y_i \uparrow X_i$ holds for all $i \in \{1,...,m\}$.
Then $Y_1,...,Y_m$ are negatively associated.
\begin{proof}
 For any disjoint subsets $A$ and $B$ of $\{1,...,m\}$, \linebreak taking $i \in A$ and $j \in B$ we have
\begin{align}
\mathbb{E}[Y_i Y_j] &= \mathbb{E} \big[ \mathbb{E}[Y_i Y_j| X_1,...,X_m] \big]  \\
&= \mathbb{E} \big[ \mathbb{E}[Y_i|X_i] \cdot \mathbb{E}[Y_j|X_j] \big]  \label{indep} \\
&\leq \mathbb{E} \big[ \mathbb{E}[Y_i|X_i] \big] \cdot \mathbb{E} \big[ \mathbb{E}[Y_j|X_j] \big] \label{neg_ass_ass} \\
& = \mathbb{E}[Y_i] \cdot \mathbb{E}[Y_j].
\end{align}
Here, (\ref{indep}) holds since each $Y_i$ only depends on $X_i$. 
\linebreak Inequality (\ref{neg_ass_ass}) follows because $X_i$ and $X_j$ are negatively associated and we 
have $\mathbb{E}[Y_i|X_i] = P(Y_i|X_i)$.
\end{proof}
\end{lemma}

\begin{lemma} \ {\bf [Chernoff]} \label{lemma:chernoff_technique} \upshape
For any real-valued random variable $X$ with finite mean $\mathbb{E}[X]$ and for any $x>0$, we have:
\begin{align}
 &DP(X,x) \leq \exp(-S(X,x)), \\
 &S(X,x) =   \sup_{\lambda} \{ \lambda x - \ln ( Z(X,\lambda) ) \}  \label{entropy},\\
 &Z(X,\lambda)=\mathbb{E}[e^{\lambda X}] \label{partition_func}.
\end{align}
The lemma follows from the observation that for $\lambda \geq 0$, we have the following
\begin{align}
P(X \geq x) =  P(e^{\lambda X} \geq e^{\lambda x}) \leq 
\inf_{\lambda} \frac{\mathbb{E}[e^{\lambda X}]}{e^{\lambda x}} . \label{exp_moment}
\end{align}
This approach is known as {\em exponential moment method} (\cite{chernoff-exp}) because of the inequality in (\ref{exp_moment}).
\end{lemma}

\begin{lemma} \ {\bf [Negative Association]} \label{lemma:negative_association} \upshape
In the exponential moment method, concentration of sums of negatively \linebreak associated random variables can 
be deduced from the \linebreak concentration of sums of their independent copies.

\begin{proof}
Let $X_1,...,X_m$ be any set of negatively \linebreak associated variables. Let $X_1',...,X_m'$ be independent shadow 
variables, i.e., independent variables such that
each $X_i'$ is distributed identically to $X_i$. Let $X=\sum_i^m X_i$ and \linebreak $X'=\sum_i^m X_i'$. 
For any set of negatively associated random \linebreak variables, one has $S(X,\epsilon) \geq S(X',\epsilon)$ since:
\begin{align}
Z(X,\lambda) &= \mathbb{E}[e^{\lambda X}] = \mathbb{E}[\prod_i^{m} e^{\lambda X_i}]& \nonumber \\
&\leq \prod_i^{m} \mathbb{E}[e^{\lambda X_i}] = \mathbb{E}[e^{\lambda X'}] = Z(X',\lambda).&
\end{align}
The lemma is due to \cite{McAllester2003} which follows from definition of 
entropy function $S$ given by (\ref{entropy}).
\end{proof}
This lemma is very helpful in the context of large deviation bounds: it implies that one can treat negatively 
 associated variables as if they were independent (\cite{McAllester2003, Dubhashi-Ranjan}).
\end{lemma}

\begin{lemma} \ {\bf [Balls and Bins]} \label{lemma:balls_and_bins} \upshape
Let $S$ be any sample \linebreak comprising $n$ items drawn i.i.d. from a fixed distribution on integers 
$\mathcal{N}=\{1,...,N\}$ (bins). 
Define $C_i$ to be the number of times that integer $i$ occurs in $S$. The random variables $C_1,...,C_N$ are negatively associated.
\begin{proof}
Let $f$ and $g$ be non-decreasing and non-increasing functions respectively. We have 
\begin{align}
\big( f(x)-f(y) \big) \big( g(x)-g(y) \big) \leq 0. \label{functional}
\end{align}
Further, assume that $X$ is a real-valued random variable and $Y$ 
is an independent shadow variable corresponding to $X$.
Exploiting (\ref{functional}), we obtain
\begin{align}
\mathbb{E}[f(X)g(X)] \leq \mathbb{E}[f(X)] \cdot \mathbb{E}[g(X)], \label{chebychev}
\end{align}
which implies that $f(X)$ and $g(X)$ are negatively \linebreak associated. Inequality (\ref{chebychev}) is an instance of 
Chebychev's \linebreak fundamental {\em association inequality}.

Now, suppose without loss of generality that $N=2$. 
Take $X \in [0,n]$, and consider the following functions
\begin{align}
\left\{
	\begin{array}{ll}
		f(X)=X, \\
		g(X)=n-X,
	\end{array}
\right.
\end{align}
where $n=C_i+C_j$ is the total counts. Since $f$ and $g$ are non-decreasing and non-increasing functions of $X$, 
\linebreak choosing $X=f(C_i)=C_i$ we have for all $i,j \in \mathcal{N}$ that
\begin{align}
\mathbb{E}[C_i \cdot C_j] \leq \mathbb{E}[C_i] \cdot \mathbb{E}[C_j],
\end{align}

which concludes the proof for $N=2$. Now, taking $f(C_i)=C_i$ and $g(C_i)=n-\sum_{j\neq i} C_j$ where \linebreak $n=\sum_{k=1}^N C_k$, 
for $N>2$ the same argument implies that $C_i$ and $C_j$ are negatively associated for all 
$i \in \mathcal{N}$ and $j \in \mathcal{N}\setminus i$.
That is to say, any increase in $C_i$ will cause a decrease in some or all of $C_j$ variables with $j \neq i$ and vice versa.
It is easy to verify that the same is true for any disjoint subsets of
the set $\{C_1,...,C_N\}$. 
\end{proof}
\end{lemma}

\begin{lemma} \ {\bf [Monotonicity]} \label{lemma:monotone} \upshape
For any negatively \linebreak associated random variables $X_1,...,X_m$ and any non-decreasing functions $f_1,...,f_m$,
we have that $f_1(X_1),...,f_m(X_m)$ are negatively associated. 
The same holds if the functions $f_1,...,f_m$ were non-increasing. 

\textbf{Remark: } The proof is in the same spirit as that of association inequality (\ref{chebychev})
and motivated by composition rules for monotonic functions that one can repeatedly apply to (\ref{functional}).
\end{lemma}

\begin{lemma} \ {\bf [Union]} \label{lemma:union} \upshape
The union of independent sets of \linebreak negatively 
associated random variables yields a set of \linebreak negatively associated random variables.

Suppose that $X$ and $Y$ are independent vectors each of which comprising a negatively associated set. 
Then, the concatenated vector $[X,Y]$ is negatively associated.
\begin{proof}
Let $[X_1, X_2]$ and $[Y_1, Y_2]$ be some arbitrary \linebreak partitions of $X$ and $Y$ respectively and assume that
$f$ and $g$ are non-decreasing functions. Then, one has
\begin{align}
\mathbb{E}[f(X_1, Y_1) \cdot  g(X_2, Y_2) ]= \nonumber \\
\mathbb{E} \big[ \mathbb{E} [f(X_1, Y_1) \cdot g(X_2, Y_2) \ | \ Y_1,Y_2] \big] \leq \nonumber \\
\mathbb{E}[\mathbb{E} [f(X_1, Y_1) \ | \ Y_1 ] \cdot \mathbb{E} [g(X_2, Y_2) \ | \ Y_2 ]] \leq \nonumber \\
\mathbb{E}[\mathbb{E} [f(X_1, Y_1) \ | \ Y_1 ]] \cdot \mathbb{E}[\mathbb{E} [g(X_2, Y_2) \ | \ Y_2 ]] = \nonumber \\
\mathbb{E}[f(X_1, Y_1)] \cdot \mathbb{E}[g(X_2, Y_2)].
\end{align}
The first inequality is due to independence of $[X_1,X_2]$ from $[Y_1, Y_2]$ which results in negative association being
preserved under conditioning and the second inequality \linebreak follows 
because $[Y_1, Y_2]$ are negatively associated (\cite{kumar_frank}). 
The same holds if $f$ and $g$ were non-increasing functions. 
\end{proof}
\end{lemma}

\begin{lemma} \ {\bf [Splitting]} \label{lemma:splitting} \upshape
Splitting an arbitrary subset of bins of any fixed discrete distribution yields a
set of negatively associated random bins.

\begin{proof}
Let $w=(w_1,...,w_m)$ be a discrete distribution and $\mathcal{W}=\{W_1,...,W_m\}$ be the associated set of random bins.
Assume that $w_i$ is split into $k$ bins $\mathcal{W}^S_i = \{ W_{i1},...,W_{ik} \}$ such 
that $w_i=\sum_{j=1}^k W_{ij}$.
Then, by Lemma \ref{lemma:balls_and_bins} members of split set $\mathcal{W}^S_i$ are negatively associated.
Clearly, the same holds for all $1 \leq i \leq m $ as well as any other 
subset of set $\mathcal{W}$. Moreover, for all $1 \leq i \leq m $ the sets 
$\mathcal{W}^S_i$ and $\mathcal{W} \setminus W_i$ are negatively associated by Lemma \ref{lemma:balls_and_bins} 
and Lemma \ref{lemma:union}. 
\end{proof}
\end{lemma}

\begin{lemma} \ {\bf [Absorption]} \label{lemma:absorption} \upshape
Absorbing any subset of bins of a discrete distribution yields negatively associated bins.
\begin{proof}
 Let $w=(w_1,...,w_N)$ be a discrete distribution and let $\mathcal{W}=\{W_1,...,W_N\}$ 
be the associated set of \linebreak random bins. Assume without loss of generality that
$\mathcal{W}^A=\{W_1^A,...,W^A_{N-1}\}$ is the absorption-induced set of random bins  
 where $w_N$ is absorbed to produce $w^A=( w^A_1,...,w^A_{N-1} )$ and  
where $w^A_i = w_i+\frac{w_N}{N-1}$ \linebreak for $i=1,...,N-1$. So, $w_N$ is discarded and we have $\sum_{i=1}^{N-1} W_i^A=1-w_N$. 
The rest of the proof concerns \linebreak applying Lemma \ref{lemma:balls_and_bins} to the absorb set $\mathcal{W}^A$. 
The same holds if we absorb $w_N$ to a subset of $\mathcal{W} \setminus W_N$ .
\end{proof}
\end{lemma}
\subsection{Negative Dependence and the Missing Mass}
For missing mass, 
the variables $W_i=\frac{C_i}{n}$ are negatively associated owing to Lemma \ref{lemma:balls_and_bins} and linearity of expectation.
Also, one has $\forall i: \ Y_i \downarrow W_i$. So, by Lemma \ref{lemma:monotone_bernoulli} 
we infer that $\forall i: \ W_i \downarrow Y_i$.
Now, $Y_1,...,Y_N$ are negatively associated because they are a set of independent binary variables with 
negative regression dependence (Lemma \ref{lemma:indep_bin_neg_reg}). Thus, concentration variables 

$Z_i=w_i Y_i - \mathbb{E}[w_i Y_i] \coloneqq \zeta(Y_i)$ are \linebreak negatively associated by Lemma \ref{lemma:monotone} 
since we have
\begin{align}
\zeta(Y_i)=
\left\{
	\begin{array}{ll}
		-w_i q_i  & \mbox{if } Y_i = 0, \\
		w_i(1 - q_i) &  \mbox{if } Y_i = 1.
	\end{array}
\right.
\end{align}
For all i, $\zeta$ is a non-decreasing function of $Y_i$.
Likewise, concentration variables $-Z_i$ are 

negatively associated.

\subsection{Information Monotonicity and Partitioning}
\begin{lemma} \ {\bf [Information Monotonicity]} \label{lemma:information_monotone} \upshape
Let \linebreak $p=(p_1,...,p_N)$ be a discrete distribution on $X=(x_1,..,x_N)$ such that for $1\leq i \leq N$ we have $P(X=x_i)=p_i$.
Suppose  we partition X into $m \leq N$ non-empty disjoint groups $G_1,...,G_m$, namely
\begin{align}
X = \cup \ G_i, \nonumber \\
\forall i \neq j: \ G_i \cap G_j = \emptyset.
\end{align}
This is called {\em coarse binning} since it generates a new distribution with groups $G_i$ whose dimensionality
is less than that of the original distribution. Note that once the distribution is transformed, 
considering any outcome $x_i$ from the original distribution we will only have access to its group membership information; 
for instance, we can observe that it belongs to $G_j$ but we will not be able to recover $p_i$.

Let us denote the induced distribution over the partition $G=(G_1,...,G_m)$ by $p^G=(p_1^G,...,p_m^G)$. Clearly, we have
\begin{align}
p_i^G=P(G_i)=\sum_{j \in G_i} P(x_j).
\end{align}
Now, consider the $f$-divergence $D_f(p^G || \ q^G)$ between \linebreak induced probability distributions $p^G$ and $q^G$.
Information monotonicity states that information is lost as we partition elements of $p$ and $q$ into groups
to produce $p^G$ and $q^G$ \linebreak respectively. Namely, for any $f$-divergence one has
\begin{align}
D_f(p^G || \ q^G) \leq D_f(p \ || \ q) \label{inf_mono},
\end{align}
which is due to Csisz\'{a}r (\cite{csiszar_survey, csiszar, amari_divergence}).
This inequality is tight if and only if for any outcome $x_i$ and partition $G_j$, we have $p(x_i | G_j) = q(x_i | G_j)$.
\end{lemma}

\begin{lemma} \ {\bf [Partitioning]} \label{lemma:partitioning} \upshape
In the exponential moment method, one can establish a deviation bound for any discrete random variable $X$ by invoking 
Chernoff's method on the associated discrete partition random variable $X^G$.

Formally, assume $X$ and $X_{\lambda}$ are discrete random variables defined on the set $\mathcal{X}$ endowed with 
probability distributions $p$ and $p_{\lambda}$ respectively.
Further, suppose that $X^G$ and $X^G_{\lambda}$ are discrete variables on a partition set $\mathcal{X}^G$
endowed with $p^G$ and $p_{\lambda}^G$ that are obtained from $p$ and $p_{\lambda}$ by partitioning using some partition $G$. 
Then, we have
\begin{align}
\forall x>0: \ DP(X, x) \leq \exp(-S(X^G,x)).
\end{align}

\begin{proof}
Let $\lambda(x)$ be the optimal $\lambda$ in (\ref{entropy}). Then, we have
\begin{align}
S(X,x) &= x \lambda(x) - \ln(Z(X,\lambda(x))) \nonumber \\ 
&= D_{KL} (p_{\lambda(x)} || \ p) \nonumber \\
& \geq D_{KL} (p_{\lambda(x)}^G || \ p^G) \nonumber \\
& = S(X^G,x), \label{entropy_ineq}
\end{align}
where we have introduced the $\lambda$-induced distribution
\begin{align}
P_{\lambda}(X=x) = \frac{e^{\lambda x} }{Z(X,\lambda)} P(X=x).
\end{align}
The inequality step in (\ref{entropy_ineq}) follows from (\ref{inf_mono}) and the \linebreak observation that
$D_{KL}$ is an instance of $f$-divergence 
where $f(v)=v \ln(v)$ with $v \geq 0$.
\end{proof}
 
\end{lemma}

\section{PROOF OF THE MAIN RESULTS}
\label{sec:proofs}
The central idea of the proof is to regulate the terms in the sum given by (\ref{eq:bernoullisum_mm}) via controlling
the magnitude of bins of the distribution using operations that preserve negative \linebreak association.
This mechanism will help defeat  the heterogeneity issue leading to the failure of standard probability 
\ inequalities described by \cite{McAllester2003}.

\subsection{Proof of Theorem 1: Upper Deviation Bound}
\label{sec:ud_missing_mass}
We consider the thresholds $\tau=\frac{\theta}{n}$ and $\tau'=\frac{2\theta}{n}$ and   
reduce the problem to one in which all bins that are larger than $\tau$ are eliminated, 
where $\theta \in \mathbb{R}$ will depend on the target deviation size $\epsilon$. 

The reduction is performed by {\em splitting} the bins that are larger than $\tau$ and then {\em absorbing} the 
bins that are smaller than $\tau$.
This is followed by choosing a threshold that yields the sharpest bound for the choice of $\epsilon$. 
It turns out that the optimal threshold will too be a function of $\epsilon$.

Let $\cI_{\tau} \subseteq \cI$ denote the subset of bins that are at most 
as large as $\tau$, $\cI_{\theta}$ the subset of bins whose magnitude is \linebreak between $\tau$ and $\tau'$, $\cI_{\tau'}$
the subset of bins larger than $\tau'$ and $\cI'_{\theta}$ and $\cI'_{\tau'}$ 
the set of bins that we obtain after splitting \linebreak members of $\cI_{\theta}$ and $\cI_{\tau'}$ respectively. 

Now, for each $i \in \cI \setminus \cI_{\tau}=\{ \cI_{\theta} \cup \cI_{\tau'} \}$ and for some $k \in \mathbb{N}$ that 
depends on $i$ (but we suppress that notation
below), we will have that $k \cdot \tau \leq w_i < (k+1) \cdot \tau$. For all such
$i$, we  define extra independent Bernoulli random variables $Y_{ij}$ with
$j \in \cJ_i := \{1, \ldots, k \}$ and their associated bins $w_{ij}$. For
$j \in \{1, \ldots, k-1\}$, $w_{ij} = \tau$ and $w_{ik} = w_i - (k-1) \cdot \tau$.
In this way, all bins that are larger than $\tau$ are split up into $k$ bins, 
each of which is in-between $\tau$ and $\tau'$; more precisely, the first $k-1$ are exactly $\tau$ and the last one may be 
larger. 
Therefore, we consider the split random variable 
$Y' = \sum_{i \in \cI_{\tau}} w_i Y_i + \sum_{i \in \{ \cI'_{\tau'} \cup \cI'_{\theta} \}} \sum_{j \in \cJ_i} w_{ij} Y_{ij}$ 
and the set $\mathcal{U}'=\{i | \ w_i < \tau' \}=\{ \cI_{\tau} \cup \cI'_{\theta} \cup \cI'_{\tau'} \}$. 
Furthermore, we introduce the random variable $Y''=\sum_{i \in \mathcal{U}''} w_i Y_i$ on the absorption-induced 
set $\mathcal{U}''=\{ i| \ \tau \leq w_i < \tau' \}$. The set $\mathcal{U}''$ is generated from $\mathcal{U}'$ as follows: we take the 
largest element $j \in \mathcal{U}'$ with $w_j < \tau$, 
update $w_l$ using \linebreak $w_l \leftarrow w_l + \frac{w_j}{|\mathcal{U}'|-1}$ for 
$\{ l \in \mathcal{U}': \ l \neq j, \ w_l < \tau \}$ and discard $w_j$. 
Repeating this procedure gives a set of bins whose sizes are in-between $\tau$ and $\tau'$ 
plus a single bin of size smaller than $\tau$; absorbing the latter into 
one of the members of the former with size $\tau$ yields $\mathcal{U}''$. 

\noindent Now, by choosing $\theta$ such that $f(\theta)= e^{-\theta}=\frac{\epsilon}{\gamma}$ and
\linebreak $\theta= f^{-1}(\frac{\epsilon}{\gamma})=\ln(\frac{\gamma}{\epsilon})$ for any $0<\epsilon<1$ and 
$e \epsilon < \gamma < e^n \epsilon$ as generic domain for $\gamma$, we derive the upper deviation 
bound for missing mass as follows
\begin{align}
&\mathbb{P}(Y  - {\mathbb E}[Y] \geq \epsilon) \leq  \\
 &\mathbb{P}(Y' - {\mathbb E}[Y] \geq \epsilon) = \label{eq:splitting}\\
 &\mathbb{P}(Y' - {\mathbb E}[Y'] + \left( {\mathbb E}[Y'] - {\mathbb E}[Y]\right) \geq \epsilon) \leq \\
  &\mathbb{P}(Y'  - {\mathbb E}[Y'] + f(\theta) \geq \epsilon)= \label{eq:compensation_u} \\
&\mathbb{P} \Big( Y' - {\mathbb E}[Y'] \geq (\frac{\gamma-1}{\gamma})\epsilon \Big) = \label{int_gamma}\\
&\exp \left(- \frac{{(\frac{\gamma-1}{\gamma})}^2 \epsilon^2}{2 (V_{\mathcal{U}''} + \frac{\alpha_u}{3} \cdot
					      (\frac{\gamma-1}{\gamma}) \cdot \epsilon)} \right) \leq &  \label{apply_bernstein_ud} \\
&\exp\left(
- \frac{{(\frac{\gamma-1}{\gamma})}^2 \epsilon^2}{2 (\color{black}\frac{\theta}{n} \color{black}\cdot \epsilon + \frac{2\theta}{3n} \cdot
					      (\frac{\gamma-1}{\gamma}) \cdot \epsilon)} \right) \leq \label{variance_bound}\\
& \inf_{\gamma} \Big\{ \exp\left(-\frac{3 n \epsilon (\gamma-1)^2}{10 \gamma^2 \ln(\frac{\gamma}{\epsilon})} \right) \Big\} 
= \label{infimization}\\
&e^{- c(\epsilon) \cdot n \epsilon}. \label{ud_linear}
\end{align}
Clearly, we will have that $\tau^*=\frac{\theta^*}{n}$ where $\theta^*=\ln(\frac{\gamma_\epsilon}{\epsilon})$.

Inequality (\ref{eq:splitting}) follows because the splitting procedure 
cannot decrease deviation probability of missing mass.

Formally, assume without loss of generality that $\cI \setminus \cI_{\tau}$ has only one element corresponding to $Y_1$, 
$\cJ_1=\{1,2\}$ and $k_1=1$ i.e. $w_1$ is split into two parts. 
Then, \linebreak deviation probability of $Y$ can be thought of as the total \linebreak probability mass associated to independent 
Bernoulli \linebreak variables $Y_1,...,Y_N$ whose weighted sum is bounded \linebreak below by some tail $t>0$. Hence, we have
\begin{align}
\mathbb{P}(Y \geq t)
&= \sum_{Y^{1N}; \ Y \geq t} P(Y_1,...,Y_N) \nonumber \\
&= \sum_{Y^{1N}; \ \mathring{Y}\geq t} \ R(Y_1) \cdot \prod_{i=2}^N R(Y_i) \nonumber \\
&+ \sum_{Y^{1N}; \ \mathring{Y}< t; \ Y \geq t} R(Y_1) \cdot \prod_{i=2}^N R(Y_i) \nonumber 
\end{align}
\begin{align}
&= \sum_{Y^{1N}; \ \mathring{Y}\geq t} \ R(Y_1) \cdot \prod_{i=2}^N R(Y_i) \nonumber \\
&+ \sum_{Y^{1N}; \ \mathring{Y}< t; \ Y \geq t, \ Y_1 = 1} R(Y_1) \cdot \prod_{i=2}^N R(Y_i) \nonumber \\
&= \sum_{Y^{2N}; \ \mathring{Y}\geq t} \prod_{i=2}^N R(Y_i) \nonumber \\
&+ \sum_{Y^{2N}; \ \mathring{Y}< t; \ Y \geq t} q_1 \cdot \prod_{i=2}^N R(Y_i), \label{last_term1}
\end{align}
where $\mathring{Y}=\sum_{i \geq 2} w_iY_i$ and $R(Y_i)=q_i$ if $Y_i=1$ and $R(Y_i)=1-q_i$ otherwise.  
\noindent Likewise, one can express the upper deviation probability of $Y'$ as follows
\footnotesize
\begin{align}
\mathbb{P}(Y' \geq t) &=\sum_{Y_{1N}; \ \mathring{Y} \geq t} \ R(Y_1) \cdot \prod_{i=2}^N R(Y_i) \nonumber \\
&+\sum_{Y_{11},Y_{12}, Y^{2N}; \ \mathring{Y} < t; \ Y' \geq t}
\Big( R(Y_{11}) \cdot R(Y_{12}) \Big) \prod_{i=2}^N R(Y_i) \nonumber \\
&=\sum_{Y_{2N}; \ \mathring{Y} \geq t} \ \prod_{i=2}^N R(Y_i) \nonumber 
\end{align}
\begin{align}
&+ \sum_{Y_{11},Y_{12}, Y^{2N}; \ \mathring{Y}< t; \ Y' \geq t}
\Big( R(Y_{11}) \cdot R(Y_{12}) \Big) \prod_{i=2}^N R(Y_i) \nonumber \\
&\geq \sum_{Y^{2N}; \ \mathring{Y} \geq t} \ \prod_{i=2}^N R(Y_i) \nonumber \\
&+\sum_{Y^{2N}; \ \mathring{Y}< t; \ Y' \geq t}
(q_{11} \cdot q_{12}) \prod_{i=2}^N R(Y_i), \label{last_term2}
\end{align}
\normalsize
where $R(Y_{ij})=q_{ij}$ if $Y_{ij}=1$ and
$R(Y_{ij})=1-q_{ij}$ otherwise.  
Thus, combining (\ref{last_term1}) and (\ref{last_term2}) we have 
\begin{align}
  &\mathbb{P}(Y' \geq t) - \mathbb{P}(Y \geq t) \geq \nonumber \\
  &\sum_{Y^{2N}; \ \mathring{Y}< t; \ Y' \geq t; \ Y \geq t} (q_{11} \cdot
  q_{12} - q_1) \prod_{i=2}^N R(Y_i)= \nonumber \\
  &\sum_{Y^{2N}; \ \mathring{Y}< t; \ Y' \geq t} (q_{11} \cdot
  q_{12} - q_1) \prod_{i=2}^N R(Y_i). \label{eq:diff_prob}
\end{align}
To complete the proof for (\ref{eq:splitting}), we require 
the expression for the difference between deviation probabilities in (\ref{eq:diff_prob}) to 
be non-negative for all \color{black}$t>0$ \color{black}which
holds if $q_1 \leq q_{11} \cdot q_{12}$. For the missing mass, this condition holds. Without loss of generality, 
assume that $w_i$ is split into two terms; namely, we have $w_i=w_{ij} + w_{ij'}$. Then, we can check the above condition as follows
\begin{align}
 q_i = (1- w_i)^n \leq (1-w_{ij})^n \cdot (1-w_{ij'})^n \nonumber \\
= {\Big( 1-\underbrace{(w_{ij}+w_{ij'})}_{w_i} + \underbrace{w_{ij} \cdot w_{ij'}}_{\geq 0} \Big)}^n. \label{split_cond}
\end{align}
One can verify using induction that (\ref{split_cond}) holds also for cases where the split operation produces more than two terms. 
Now, choosing tail size $t=\epsilon+\mathbb{E}Y$ implies (\ref{eq:splitting}). 

\noindent Inequality (\ref{eq:compensation_u}) follows because the gap between the expectations will be negligible.
Denoting $\mathbb{E} [Y'_i] = q'_i$, we have
\begin{align}
q'_i=
\left\{
	\begin{array}{lll}
		q_i  & \mbox{if } i \in \cI_{\tau}, \\
		q_{ij} &  \mbox{if } i \in \{ \cI'_{\tau'} \cup \cI'_{\theta} \}, \\
		0 & \mbox{otherwise. }
	\end{array}
\right.
\end{align}
Namely, we can write
\begin{align}
g_u(\theta) &= {\mathbb E}[Y'] - {\mathbb E}[Y] = \sum_{i \in \cI} w_i (q'_i-q_i) \nonumber \\
& =\sum_{i \in  \cI_{\tau}} w_i q_i + 
 \sum_{i \in \{ \cI'_{\tau'} \cup \cI'_{\theta} \}} \sum_{j \in \cJ_i} w_{ij} q_{ij} - \sum_{i \in \cI} w_i q_i \nonumber \\ 
&=\sum_{i \in \{ \cI'_{\tau'} \cup \cI'_{\theta} \}} \sum_{j \in \cJ_i} w_{ij} q_{ij} - \sum_{i \in \{ \cI_{\tau'} \cup \cI_{\theta} \}} w_i q_i \nonumber \\
&\leq \sum_{i \in \{ \cI'_{\tau'} \cup \cI'_{\theta} \}} \sum_{j \in \cJ_i} w_{ij} q_{ij} \nonumber \\
&\leq \sum_{i \in \{ \cI'_{\tau'} \cup \cI'_{\theta} \}} \sum_{j \in \cJ_i} w_{ij} f(\theta) \leq f(\theta). \label{compensation_u}
\end{align}

The expression in (\ref{apply_bernstein_ud}) is Bernstein's inequality applied to the random variable 
$Z_u = \sum_{i \in \mathcal{U}''}  Z_i$ relying upon \linebreak Lemma \ref{lemma:partitioning}.
Here, the concentration variables are \linebreak 
$Z_i =  w_i Y_i - \mathbb{E}[w_iY_i]$ with $\ i \in \mathcal{U}''$ and we set $\alpha_u=\tau'$. 

Let $V_{\mathcal{U}''}$ be variance proxy term $V$ in Bernstein's inequality as defined in (\ref{eq:bernsteina}) 
attached to $\mathcal{U}''$. The functions $f, g: (0,1) \times \mathbb{N} \rightarrow (0,1)$ with $f(x,n)=x(1-x)^n(1-(1-x)^n)$ 
and $g(x,n)=x^2(1-x)^n(1-(1-x)^n)$ are non-increasing with respect to $x$ on $(\frac{1}{n+1},1)$ and $(\frac{2}{n+2},1)$ respectively.
We obtain for $1< \theta < n$, an upperbound on $V_{\mathcal{U}''}$ as follows:
\begin{align}
V_{\mathcal{U}''}
&=\sum_{i: \ w_i \in \mathcal{U}''} w_i^2 (1-w_i)^n \Big(1-{(1-w_i)}^n \Big) \nonumber \\
& \leq \tau \cdot \sum_{i: \ w_i \in \mathcal{U}''} w_i (1-w_i)^n \Big(1-{(1-w_i)}^n \Big) \nonumber \\
& = \tau \cdot {\underline{\sigma}}^2_{\mathcal{U}''}  \nonumber \\
& \leq \tau \cdot \sum_{i: \ \tau \leq w_i < \tau'; \ \sum_i w_i=1} w_i (1-w_i)^n \nonumber \\
& \leq \underbrace{|\cI_{(\theta,n)}|}_{\leq \frac{n}{\theta}} \cdot \big(\frac{\theta}{n}\big)^2 \cdot 
\Big( 1-\frac{\theta}{n}\Big)^n \nonumber \\
& \leq \frac{\theta}{n} \cdot e^{-\theta} < \frac{\theta}{n} \cdot \epsilon.
\end{align}

\noindent In order to see why (\ref{ud_linear}) holds, consider
$c(\gamma, \epsilon)=\frac{\epsilon (\gamma-1)^2}{\gamma^2 \ln(\frac{\gamma}{\epsilon})}$ and let us examine the derivatives as follows
\begin{align}
\frac{\partial c(\gamma, \epsilon)}{\partial \gamma}= -\frac{\epsilon^2 (\gamma-1) (\gamma -1 -2 \ln{(\frac{\gamma}{\epsilon}}))}
{\gamma^3 \ln^2{(\frac{\gamma}{\epsilon}})}, \label{dc_dgam} \\
\frac{\partial^2 c(\gamma, \epsilon)}{{\partial \gamma}^2} =
\frac{{\epsilon}^2}{\gamma^4 \ln^3{(\frac{\gamma}{\epsilon})}} 
\Big[ (6-4\gamma) \ln^2{(\frac{\gamma}{\epsilon}}) + \nonumber \\
(\gamma^2-6\gamma+5)\ln{(\frac{\gamma}{\epsilon}}) + 2(\gamma-1)^2 \Big]
. \label{dc_d2gam}
\end{align}
Solving for the first derivative using (\ref{dc_dgam}), we obtain
\begin{align}
\gamma_{\epsilon} = -2W_{-1} \Big(-\frac{\epsilon}{2 \sqrt e} \Big). \label{lampert_w}
\end{align}

Inspecting the second derivative given by (\ref{dc_d2gam}), we can see that the function
$c(\gamma, \epsilon)$ is concave with respect to $\gamma$ for any $\gamma>2$. Recall, moreover, that there are
interrelated restrictions on $\gamma$, $\epsilon$ and $n$ in derivation of (\ref{infimization}) and (\ref{ud_linear})
which are collectively expressed as
\begin{align}
\text{max} \{ e \cdot \epsilon , \ 1, \ 2, {\gamma}(1) \} <\gamma< e^n, \quad n \geq \lceil {\gamma}_{\epsilon} \rceil -1. \label{domain_res}
\end{align}

\subsection{Proof of Theorem 2: Lower Deviation Bound}
\label{sec:ld_missing_mass}

The proof for lower deviation bound proceeds in the same spirit as section \ref{sec:ud_missing_mass}. The idea is 
again to reduce the problem to one in which all bins that are larger than the threshold $\tau$ are eliminated.

We split large bins and then absorb small bins to enable us shrink the variance while controlling
the magnitude of terms (and consequently the key quantities $\alpha$ and $V$) before applying Bernstein's inequality.

By choosing $\theta$ such that $f(\theta)= e^{-\theta}$ so that $\theta=\ln(\frac{\gamma}{\epsilon})$, 
for any $0<\epsilon<1$ with $ e \epsilon < \gamma < e^n \epsilon$ being generic domain for $\gamma$ 
we obtain a lower deviation bound 
as follows
\begin{align}
&\mathbb{P}(Y  - {\mathbb E}[Y] \leq -\epsilon) \leq& \\
& \mathbb{P}(Y' - {\mathbb E}[Y] \leq -\epsilon) = \\
& \mathbb{P}(Y' - {\mathbb E}[Y'] + \left( {\mathbb E}[Y'] - {\mathbb E}[Y]\right) \leq -\epsilon) \leq \\
& \mathbb{P}(Y'  - {\mathbb E}[Y'] - f(\theta) \leq -\epsilon) = \label{eq:compensation_l} \\
&\mathbb{P} \Big( Y'-\mathbb{E}[Y'] \leq -(\frac{\gamma-1}{\gamma})\epsilon \Big) \color{black}\leq  \label{introduce_gamma4} \\
&\leq \exp\left(- \frac{{(\frac{\gamma-1}{\gamma})}^2 \epsilon^2}{2 (V_{\mathcal{L}''} + \frac{\alpha_l}{3} \cdot
					(\frac{\gamma-1}{\gamma}) \cdot \epsilon)} \right) \leq & \label{apply_bernstein_ld} \\
&\leq \exp\left(- \frac{{(\frac{\gamma-1}{\gamma})}^2 \epsilon^2}{2 (\frac{\theta}{n} \cdot \epsilon + \frac{2\theta}{3n} \cdot
(\frac{\gamma-1}{\gamma}) \cdot \epsilon)} \right) \leq & \label{variance_bound_l} \\
&\inf_{\gamma} \Big\{
\exp \Big(- \frac{3 n \epsilon (\gamma-1)^2}{10 \gamma^2 \ln(\frac{\gamma}{\epsilon})} \Big) \Big\} =& \label{phi_func_l}\\
&e^{- c(\epsilon) \cdot n \epsilon}, & \label{ld_linear}
\end{align}

where $c(\epsilon)$ and $\tau^*$ are as before and domain restrictions are determined similar to (\ref{domain_res}). 

The variables $Y'$ and $Y''$, and the sets $\mathcal{L}'$ and $\mathcal{L}''$ are defined in the same fashion 
as Section \ref{sec:ud_missing_mass}.

The first inequality is proved in the same way as (\ref{eq:splitting}).
Now, we set $\mathbb{E} [Y'_i] = q'_i$ such that
\begin{align}
q'_i=
\left\{
	\begin{array}{lll}
		q_i  & \mbox{if } w_i < \tau', \\
		0 & \mbox{otherwise. }
	\end{array}
\right. 
\end{align}
Inequality (\ref{eq:compensation_l}) follows because the compensation gap will remain small since we have
\begin{align}
g_l(\theta) &= {\mathbb E}[Y'] - {\mathbb E}[Y] = \sum_{i \in \cI} w_i (q'_i-q_i) \nonumber \\
&= \sum_{i: w_i < \tau'} w_i q_i - \sum_{i \in \cI} w_i q_i
= -\sum_{i: w_i \geq \tau'} w_i q_i \nonumber \\
&\geq -\sum_{i: w_i \geq \tau'} w_i f(\theta) \geq -f(\theta).
\end{align}

The expression given by (\ref{apply_bernstein_ld}) is Bernstein's inequality \linebreak applied to random variable 
$Z_l = \sum_{i \in \mathcal{L}''} Z_i$ where we have defined  
$Z_i =  w_i (\mu - w_i Y_i) - \mathbb{E}[w_i(\mu - w_i Y_i)]$ with $\mu$ being 
the upper bound on the value of the $w_i Y_i$ terms. 

Further, we choose $\alpha_l=\tau'$.
Observe that $Z_l=-Z_u$ and $\mu=\alpha_l$. Finally, an upperbound on $V_{\mathcal{L}''}$ can 
be determined with arguments identical to that of $V_{\mathcal{U}''}$. 

The rest of the proof proceeds in an analogous manner to the proof of upper deviation bound.

\section{CONCLUSIONS}
\label{sec:conclusion}
We proposed a new technique for establishing concentration inequalities and applied it to the 
missing mass using Bernstein's inequality. Along the way, we introduced a \linebreak collection of concepts and tools in 
the intersection of probability theory and information theory that have the potential to be advantageous in more general settings. 

Recall that Bernstein's inequality hinges on establishing an upperbound on $Z(X,\lambda)$ given 
by (\ref{partition_func}) in a particular way. Clearly, this choice is not unique and one can choose 
any other upperbound (e.g. c.f. \cite{Lugosi_Concentration}) and apply the same technique to derive 
potentially tight bounds achievable within the framework of exponential moment method.

Our bounds sharpen the leading results for missing mass in the case of small deviations. These  
 inequalities hold subject to the mild condition that the sample size is large enough, namely 
$n \geq \lceil {\gamma}_{\epsilon} \rceil -1$.

We select the best known bounds in Berend and \linebreak Kontorovich (2013) 
 for the comparison.
Our lower deviation and upper deviation bounds improve state-of-the-art for any $0<\epsilon<0.021$ and 
any $0<\epsilon<0.045$ respectively.

Plugging in the definitions, we can see that the \linebreak compensation gap can be expressed as a function of $\epsilon$
and show that the following holds 
\begin{align}
|g(\epsilon)| \leq \sqrt e \cdot \exp \Big( W_{-1}(\frac{-\epsilon}{2 \sqrt e}) \Big), \label{gap_eps}
\end{align}
where we have dropped the subscript of gap $g$.
Note that the gap is negligible for small $\epsilon$ compared to large values of $\epsilon$ for both
(\ref{ud_linear}) and (\ref{ld_linear}). This observation supports the
fact that we obtained sharper bounds for small deviations. 

Mathematical analysis of missing mass via concentration inequalities has various important
 applications including density  estimation, generalization bounds and handling missing data just to name a few. 
Needless to say that any refinement in bounds or tools developed for the former may directly contribute to 
advancement in those applications.

\section*{Acknowledgements}
\label{sec:Acknowledgement}
The authors are grateful to National ICT Australia (NICTA) for generous funding, 
as part of collaborative machine learning research projects. We would like to thank \linebreak 
Peter Gr\"{u}nwald, Aryeh Kontorovich, Thijs van Ommen and Mark Schmidt for feedback and helpful discussions, 
and anonymous reviewers for their constructive comments.

\bibliography{missing_mass}
\bibliographystyle{plainnat}
\end{document}